\documentclass[runningheads,a4paper]{llncs}

\usepackage{color}
\usepackage{authblk}

\usepackage{etoolbox}

\newcommand{\repthanks}[1]{\textsuperscript{\ref{#1}}}
\makeatletter
\patchcmd{\maketitle}
  {\def\thanks}
  {\let\repthanks\repthanksunskip\def\thanks}
  {}{}
\patchcmd{\@maketitle}
  {\def\thanks}
  {\let\repthanks\@gobble\def\thanks}
  {}{}
\newcommand\repthanksunskip[1]{\unskip{}}
\makeatother

\usepackage{amssymb}
\usepackage{graphicx}

\usepackage{url}
\urldef{\mailsa}\path|{alfred.hofmann, ursula.barth, ingrid.haas, frank.holzwarth,|
\urldef{\mailsb}\path|anna.kramer, leonie.kunz, christine.reiss, nicole.sator,|
\urldef{\mailsc}\path|erika.siebert-cole, peter.strasser, lncs}@warwick.ac.uk|    
\newcommand{\keywords}[1]{\par\addvspace\baselineskip
\noindent\keywordname\enspace\ignorespaces#1}

\begin{document}

\mainmatter  % start of an individual contribution

% first the title is needed
\title{Context-Aware Learning using Transferable Features for Classification of Breast Cancer Histology Images}

\author[1]{Ruqayya Awan\thanks{Joint co-authors\protect\label{X}}}
\author[1,2]{Navid Alemi Koohbanani\repthanks{X}}
\author[1]{Muhammad Shaban}
\author[1]{Anna Lisowska}
\author[1,2,3]{Nasir Rajpoot}
\affil[1]{Department of Computer Science, University of Warwick, Coventry, UK}
\affil[2]{The Alan Turing Institute, London, UK}
\affil[3]{Department of Pathology, University Hospitals Coventry \& Warwickshire, UK}

% a short form should be given in case it is too long for the running head
\titlerunning{BACH Challenge}
\institute{}

% the name(s) of the author(s) follow(s) next
%
% NB: Chinese authors should write their first names(s) in front of
% their surnames. This ensures that the names appear correctly in
% the running heads and the author index.
%

%\authorrunning{In Proceedings of ICIAR 2018}
\authorrunning{BACH Challenge}

%
% NB: a more complex sample for affiliations and the mapping to the
% corresponding authors can be found in the file "llncs.dem"
% (search for the string "\mainmatter" where a contribution starts).
% "llncs.dem" accompanies the document class "llncs.cls".
%

\toctitle{Lecture Notes in Computer Science}
\tocauthor{Authors' Instructions}
\maketitle

\begin{abstract}
Convolutional neural networks (CNNs) have been recently used for a variety of histology image analysis. However, availability of a large dataset is a major prerequisite for training a CNN which limits its use by the computational pathology community. In previous studies, CNNs have demonstrated their potential in terms of feature generalizability and transferability accompanied with better performance. Considering these traits of CNN, we propose a simple yet effective method which leverages the strengths of CNN combined with the advantages of including contextual information, particularly designed for a small dataset. Our method consists of two main steps: first it uses the activation features of CNN trained for a patch-based classification and then it trains a separate classifier using features of overlapping patches to perform image-based classification using the contextual information. The proposed framework outperformed the state-of-the-art method for breast cancer classification.

\end{abstract}

\section{Introduction}

Breast cancer is the most common type of cancer diagnosed and is the second most common type of cancer with high mortality rate after lung cancer in women \cite{siegel2016cancer}. Due to the increased incidence of breast cancer and subjectivity in diagnosis, there is an increasing demand for automated systems. To this end, deep neural networks (DNNs) have been widely used to produce the state-of-the-art results for a variety of histology image analysis tasks such as nuclei detection and classification \cite{sirinukunwattana2016locality}, tissue classification \cite{cruz2014automatic,wang2016deep} and segmentation \cite{chen2016dcan,bejnordi2017diagnostic}. 

The CAMELYON16 challenge \cite{bejnordi2017diagnostic} is the best demonstration of using deep learning for automatic tissue analysis, outperforming the pathologists in terms of detection of tumors within the whole slide images (WSIs). The objective of this challenge was to automatically detect the metastasis in haematoxylin and eosin (H\&E) stained WSIs of lymph node sections. Cruz-Roa \emph{et al.} \cite{cruz2014automatic} presented a deep learning architecture for automated basal carcinoma detection. This method first learns image representation via autoencoder and then a CNN is applied on this representation to capture both translation invariant features and a compact image representation. Spanhol \emph{et al.} \cite{spanhol2016breast} applied a simple CNN for classifying the BreaKHis database \cite{spanhol2016dataset} consisting of microscopic images of benign and malignant breast tumor biopsies. Small patches were extracted at different magnification levels to train the network and during inference, final output was produced by combining the predictions of the small patches.

%\subsection{Transferable Features}

The generalizability property of DNN makes their features transferable to other applications which encouraged the researchers to employ transfer learning for histology images as in \cite{chen2016dcan,bayramoglu2016transfer,han2017breast}. These features have also been used to train separate classifiers for predictions \cite{xu2017large,valkonen2017dual,xu2015deep,araujo2017classification}, which are particularly useful when there is not enough dataset for training the CNN from scratch. In some recent studies \cite{agarwalla2017representation,bejnordi2017context}, context-aware based learning architecture has been introduced, in which first CNN is trained using high pixel resolution patches to extract features at a cellular level that are then fed to a second CNN, stacked on top of the first for expanding the context from a single patch to a large tissue region. The experimental results of these studies suggest that the contextual information plays a crucial role in identifying abnormalities in heterogeneous tissue structures.

%\subsection{Our Contribution}

Our contribution in this work is twofold. First, we propose to use CNN features as a generic descriptor for a small dataset, provided as a part of a challenge dataset. We extract transferable features from a number of networks, each trained on a different dataset for the purpose of classification by a separate classifier trained on these features. As our second contribution, we combine these features to learn context of a large patch to improve our classification performance. To this end, we use transferable features for a block of consecutive patches to train a SVM model to classify the H\&E stained breast images into normal, benign, carcinoma \emph{insitu} (CIS) and breast invasive carcinoma (BIC). 

%\section{Literature Review}

\section{Dataset and Experimental Setup}

We used the dataset provided as a part of the ICIAR 2018 challenge for the classification of breast cancer histology images. This dataset comprises of 400 high resolution images of size 2048$\times$1536 pixels at 20$\times$ magnification, stained with H\&E stain. The pixel resolution for these images is 0.42 $\mu$m. Each image belongs to one of the four classes: normal, benign, \emph{insitu} carcinoma or invasive carcinoma. The ground truth was provided by the two pathologists. To study the feature transferability of CNN, we experimented with other part of the challenge dataset provided for segmentation task. Ten WSIs with coarse annotations were provided for this task. We extracted patches from these WSIs after manually refining the original annotations.

The challenge dataset for a classification task consists of training images used in \cite{araujo2017classification} along with 151 additional images. To evaluate the effectiveness of our proposed approach, we splitted the challenge dataset for two settings. In the first setting, we use the same images for training and testing which were used in \cite{araujo2017classification} for a fair comparison. We included the additional images in our validation set while training the network. The test dataset contains two set of images, with equal number of images in each class. The testing data is not provided with the challenge data but is made publicly available by the authors in two sets. The first test set contains 20 images while the second set contains 16 images and is referred as ’test extended’ dataset in this paper. In the second setting which is used for submission to the challenge, we combined the whole challenge dataset from task-1 and the test dataset and randomly split them into 75\% training and 25\% validation set.

Regarding the implementation, we used residual neural network with 50 layers for patch-based classification in Tensorflow. For context-aware image-based classification, support vector machine (SVM) classifier with radial basis function (RBF) was used and implemented in MATLAB. Further details on both these steps are given in the \textit{Methods} section.

\section{Methods}

\begin{figure}[ht]
\centering
\includegraphics[width = \linewidth]{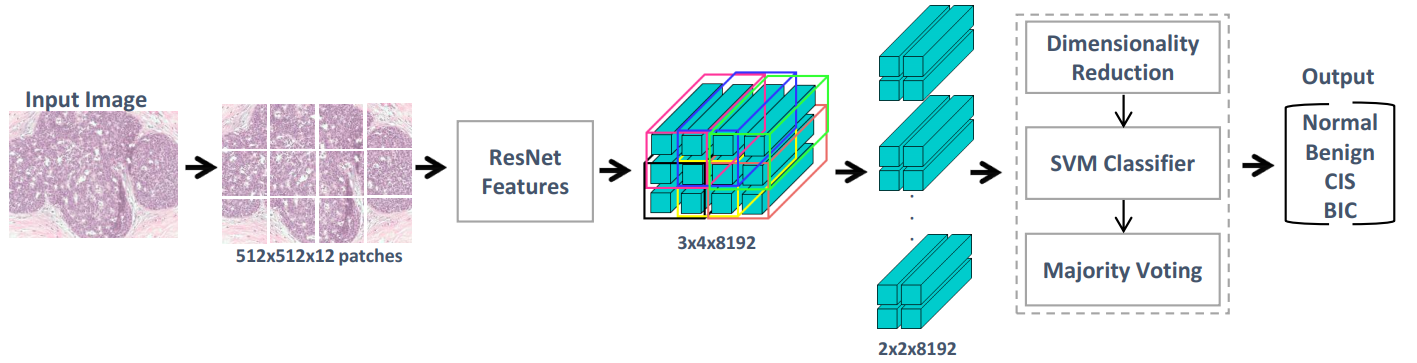}
%\vspace*{-8mm}
\caption{Flow Diagram of our classification framework. Twelve non-overlapping patches are extracted from the input image. A 8192-dimensional feature vector is then obtained for each patch using a trained ResNet. The class label for the overlapping blocks (2$\times$2) of features is obtained from the SVM model after reducing the dimensionality. The final image-based prediction is made using the majority voting approach.   }
\label{fig:system_overview}
\end{figure}

In this paper, we introduce an effective model for the purpose of image-based classification using more context information, particularly for a small dataset. To this end, we design our model in two main steps: patch-based classification and context-aware image-based classification. The overall system architecture is shown in Figure \ref{fig:system_overview}.

\subsection{Preprocessing}

Stain inconsistency of digitized WSIs is a significant issue affecting the performance of machine learning (ML) systems. The dataset provided for this challenge contains images with large stain variation. To this end, we performed stain normalization using the Reinhard method \cite{reinhard2001color}, available in our group's \textit{Stain Normalization Toolbox} \cite{khan2014nonlinear}. This method transforms the color distribution of an image to the color distribution of a target image by matching the mean and standard deviation of the source image to that of target image. This transformation is carried out for each channel separately, in the Lab colorspace.

\begin{figure}[ht]
\centering
\includegraphics[width = \linewidth]{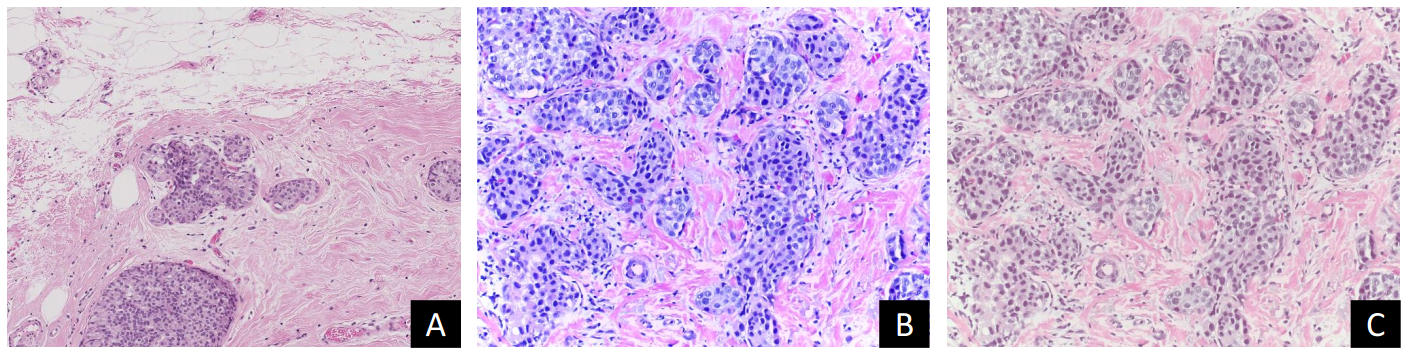}
%\vspace*{-60mm}
\caption{Output of stain normalization: A, B and C show the target image, the original image and the stain normalized version of B respectively.}
\label{fig:example}
\end{figure}

\subsection{Patch-based classification}

ResNet \cite{he2016deep} introduced in 2015 by Microsoft has been shown to outperform several architectures including VGG \cite{simonyan2014very}, GoogleNet \cite{szegedy2015going}, PReLU-net \cite{he2015delving} and BN-inception \cite{ioffe2015batch}. This network also outperformed best performing networks with a significant margin for the classification of histopathology colorectal images \cite{korbar2017deep}. The state-of-the-art results of ResNet on different datasets motivated us to use it for our patch-based classification. For our experiments, we used ResNet with 50 layers. For network training, overlapping patches of size 512$\times$512 pixels were extracted from the images. The network was trained for 16 epochs with batch size of 12 and the best trained network was selected for further processing. The training was done using stochastic gradient descent with momentum set to 0.95. The learning rate was initially set to 0.001 and was decremented after each update. Due to the very small dataset and also to make our network robust to feature transformation, we performed data augmentation involving random rotation (90 to 360 degrees with step of 90 degrees) and flipping during the training stage.
 
\subsection{Context-aware image-wise classification}    

The above patch-based classification network learns a limited contextual representation for each class by using small patches of size 512$\times$512 pixels. To train a classifier with larger context, we divided each image into twelve non-overlapping patches and for each patch, we then extracted 8192 dimensional feature vector from the last layer of our patch-based network. We then trained an SVM classifier with the flattened features of 2$\times$2 overlapping blocks of patches which is equivalent to training the classifier with the features of patch 1024$\times$1024 pixels in size. Principal component analysis (PCA) was performed to reduce the dimensionality. The image level label was then decided by majority voting on the labels of overlapping blocks. This approach has been shown to improve the results in previous studies by training a CNN using the features of bigger blocks of patches. Due to the limited availability of dataset for training, we chose to train an SVM classifier rather than a CNN. To increase the number of samples for SVM training, we augmented the training samples using rotation and flipping.
 
\section{Experimental Results}

For the evaluation of our proposed method, we experimented with different configurations to show the significance of contextual information, effect of feature transferability using networks trained on different datasets and also to compare our method with the results of \cite{araujo2017classification}.

\begin{figure}[ht]
\centering
\includegraphics[width = 10cm]{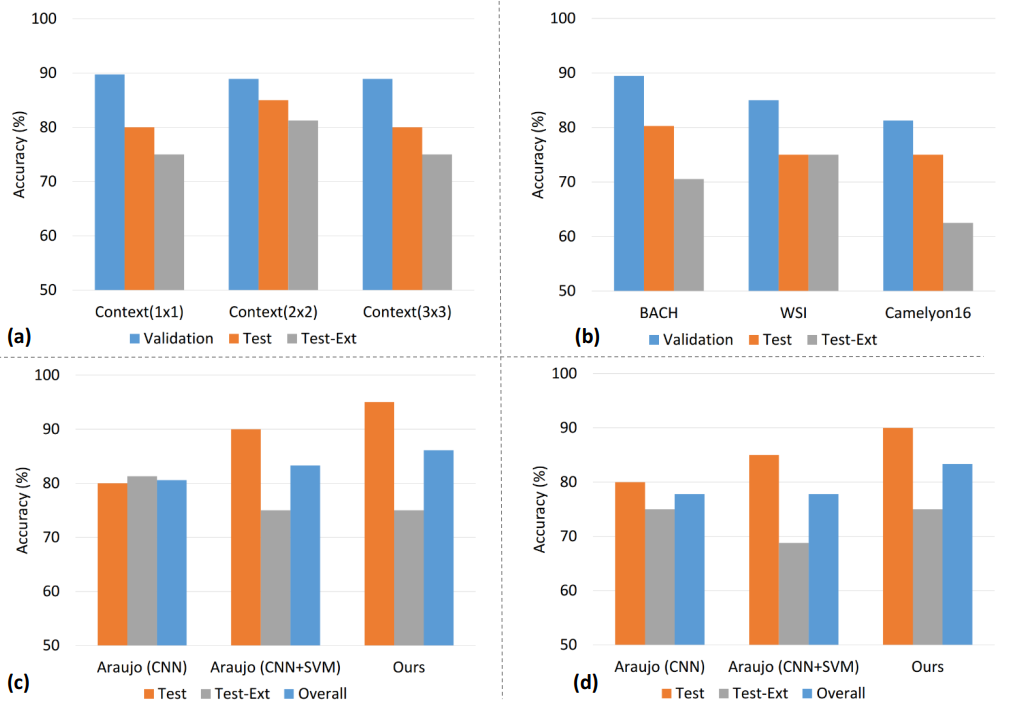}
%\vspace*{-4mm}
\caption{Summary of our experimental results. (a) Accuracy obtained using the context of various size of blocks where Context(1$\times$1), Context(2$\times$2) and Context(3$\times$3) represent contextual block of size 512$\times$512, 1024$\times$1024 and 1536$\times$1537 pixels. (b) Analysis of transferable features using three networks, each trained with a different dataset where BACH, WSI and Camelyon16 represent challenge dataset for classification, challenge dataset for segmentation and Camelyon16 challenge dataset respectively. (c) and (d) show comparative analysis of our approach with the previous study \cite{araujo2017classification} using both 2-class and 4-class classification respectively.}
\label{fig:comparison_results}
\end{figure}

Firstly, we experimented with contextual information captured from the varying size of block of patches. We trained SVM with the context of 1$\times$1 block (512$\times$512 pixels), 2$\times$2 block (1024$\times$1024 pixels) and 3$\times$3 block (1536$\times$1536 pixels) of patches. As shown in Figure \ref{fig:comparison_results} (a), we obtained better accuracy with 2$\times$2 block as compared to the 1$\times$1 block but we observed a decline in accuracy for 3$\times$3 block of context. It is due to the fact that the increase in context using the large block of patches reduces the amount of data for training SVM. Otherwise, incorporating a larger context could have improved our results if we had enough dataset. For our further experiments, we used 2$\times$2 block of context for classification.

To study the feature transferability, we used activation features from three networks, each trained on a different dataset: training images used in \cite{araujo2017classification}, challenge dataset (WSIs) for segmentation task and the Camelyon dataset \cite{bejnordi2017diagnostic}. To enrich the SVM training, we added half of the additional challenge dataset to our training set. Our results demonstrate that the features learned by the first network tend to provide the most appropriate representation for training the SVM. The network trained on the WSIs has been shown to be the second best feature descriptor as the dataset used to train this network is similar to the evaluation data. While our third network, since it is trained on a completely different dataset did not perform well as compared to the other networks.
  
For comparative analysis of our proposed method, we compared our results with the results of previously published work of the organizers of BACH challenge \cite{araujo2017classification}. For a fair comparison, we used the same dataset as in \cite{araujo2017classification} for training and testing. Along with 4-class classification, we also present comparison using 2-class classification, for which we grouped normal and benign images in one class while CIS and BIC were grouped in the other class. The 2-class and 4-class comparative results are shown in Figure \ref{fig:comparison_results} (c) and (d) respectively. Our method achieved higher accuracy compared to \cite{araujo2017classification} which demonstrates the capability of the contextual information for discriminating different classes.

\section{Conclusion}

In this paper, we proposed a context-aware network for automated classification of breast cancer histology images. The proposed method leverages the power of CNNs to encode the representation of a patch into high dimensional space and uses traditional machine method (SVM) to aggregate the contextual information from the high dimensional features while having a limited dataset. Our proposed approach outperformed the existing methods proposed for the same task. The proposed method is not limited to breast cancer classification task. It could be applied to other problems where both high resolution and contextual information are required to make an optimal prediction.

%\bibliography{references}
%\bibliographystyle{ieeetr}		%\bibliographystyle{plain}

\end{document}